\documentclass[11pt,a4paper]{article}
\usepackage[hyperref]{naaclhlt2018}
\usepackage{times}
\usepackage{latexsym}
\usepackage{adjustbox}
\usepackage{multirow}
\usepackage{float}
\usepackage{hyperref}
\usepackage{url}

\aclfinalcopy % Uncomment this line for the final submission
\newcommand\smalltt[1]{\texttt{\small #1}}
\newcommand{\specialcell}[2][l]{%
  \begin{tabular}[#1]{@{}l@{}}#2\end{tabular}}

\title{Integrating Multiplicative Features\\into Supervised Distributional Methods for Lexical Entailment}

\author{Tu Vu \\
  College of Information and Computer Sciences  \\
  University of Massachusetts Amherst \\
  Amherst, MA, USA  \\
  {\tt tuvu@cs.umass.edu} \\\And
  Vered Shwartz \\
  Computer Science Department \\
  Bar-Ilan University \\
  Ramat-Gan, Israel \\
  {\tt vered1986@gmail.com} \\}

\date{}

\begin{document}
 \maketitle

\begin{abstract}
Supervised distributional methods are applied successfully in lexical entailment, but recent work questioned whether these methods actually learn a relation between two words. Specifically, \newcite{Levy:15} claimed that linear classifiers learn only separate properties of each word. We suggest a cheap and easy way to boost the performance of these methods by integrating multiplicative features into commonly used representations. We provide an extensive evaluation with different classifiers and evaluation setups, and suggest a suitable evaluation setup for the task, eliminating biases existing in previous ones.
\end{abstract}

\section{Introduction}
\label{sec:intro}
Lexical entailment is concerned with identifying the semantic relation, if any, holding between two words, as in \emph{(pigeon, hyponym, animal)}.
The popularity of the task stems from its potential relevance to various NLP applications, such as question answering and recognizing textual entailment \cite{dagan2013recognizing} that often rely on lexical semantic resources with limited coverage like Wordnet \cite{Miller:95}. Relation classifiers can be used either within applications or as an intermediate step in the construction of lexical resources which is often expensive and time-consuming.

Most methods for lexical entailment are distributional, i.e., the semantic relation holding between $x$ and $y$ is recognized based on their distributional vector representations. While the first methods were unsupervised and used high-dimensional sparse vectors \cite{W03-1011,Kotlerman:10,Santus:14}, in recent years, supervised methods became popular \cite{Baroni:12,Roller:14,Weeds:14}. These methods are mostly based on word embeddings \cite{mikolov2013distributed,pennington2014glove} utilizing various vector combinations 
%, e.g., concatenation \cite{Baroni:12} 
%or difference \cite{Weeds:14}
that are designed to capture relational information between two words. 

While most previous work reported success using supervised methods, some questions remain unanswered: First, several works suggested that supervised distributional methods are incapable of inferring the relationship between two words, but rather rely on independent properties of each word \cite{Levy:15,Roller:16,Shwartz:16a}, making them sensitive to training data; Second, it remains unclear what is the most appropriate representation and classifier; previous studies reported inconsistent results with \textbf{Concat}$\langle \vec{v_x} \oplus \vec{v_y} \rangle$ \cite{Baroni:12} and \textbf{Diff}$\langle \vec{v_y} - \vec{v_x} \rangle$ \cite{Roller:14,Weeds:14,Fu:14}, using various classifiers.

In this paper, we investigate the effectiveness of multiplicative features, namely, the element-wise multiplication \textbf{Mult}$\langle \vec{v_x} \odot \vec{v_y} \rangle$, and the squared difference \textbf{Sqdiff}$\langle (\vec{v_y} - \vec{v_x}) \odot (\vec{v_y} - \vec{v_x}) \rangle$. These features, similar to the cosine similarity and the Euclidean distance, might capture a different notion of interaction information about the relationship holding between two words. We directly integrate them into some commonly used representations. For instance, we consider the concatenation $\textbf{Diff} \oplus \textbf{Mult}$ $\langle (\vec{v_y} - \vec{v_x}) \oplus (\vec{v_x} \odot \vec{v_y})\rangle$ that might capture both the typicality of each word in the relation (e.g., if $y$ is a typical hypernym) and the similarity between the words.

We experiment with multiple supervised distributional methods and analyze which representations perform well in various evaluation setups. Our analysis confirms that integrating multiplicative features into standard representations can substantially boost the performance of linear classifiers. While the contribution over non-linear classifiers is sometimes marginal, they are expensive to train, and linear classifiers can achieve the same effect ``cheaply'' by integrating multiplicative features. The contribution of multiplicative features is mostly prominent in strict evaluation settings, i.e., lexical split \cite{Levy:15} and out-of-domain evaluation that disable the models' ability to achieve good performance by memorizing words seen during training. We find that \textbf{Concat} $\oplus$ \textbf{Mult} performs consistently well, and suggest it as a strong baseline for future research.

\section{Related Work}
\label{sec:background}
\paragraph{Available Representations}
In supervised distributional methods, a pair of words $(x,y)$ is represented as some combination of the word embeddings of $x$ and $y$, most commonly \textbf{Concat} $\langle \vec{v}_x \oplus \vec{v}_y \rangle$ \cite{Baroni:12} or \textbf{Diff} $\langle \vec{v}_y - \vec{v}_x \rangle$ \cite{Weeds:14,Fu:14}. 

\paragraph{Limitations}
Recent work questioned whether supervised distributional methods actually learn the relation between $x$ and $y$ or only separate properties of each word. \newcite{Levy:15} claimed that they tend to perform ``lexical memorization'', i.e., memorizing that some words are prototypical to certain relations (e.g., that $y=$ \emph{animal} is a hypernym, regardless of $x$). \newcite{Roller:16} found that under certain conditions, these methods actively learn to infer hypernyms based on separate occurrences of $x$ and $y$ in Hearst patterns \cite{Hearst:92}. In either case, they only learn whether $x$ and $y$ independently match their corresponding slots in the relation, a limitation which makes them sensitive to the training data \cite{Shwartz:17,Sanchez:17}.

\paragraph{Non-linearity}
\newcite{Levy:15} claimed that the linear nature of most supervised methods limits their ability to capture the relation between words. They suggested that using support vector machine (SVM) with non-linear kernels slightly mitigates this issue, and proposed \smalltt{KSIM}, a custom kernel with multiplicative integration. 

\paragraph{Multiplicative Features}
The element-wise multiplication has been studied by \newcite{Weeds:14}, but models that operate exclusively on it were  not competitive to \textbf{Concat} and \textbf{Diff} on most tasks.  \newcite{Roller:14} found that the squared difference, in combination with \textbf{Diff}, is useful for hypernymy detection. Nevertheless, little to no work has focused on investigating 
combinations of representations obtained by concatenating various base representations for the more general task of lexical entailment. 

\section{Methodology}
\label{sec:methodology}
\begin{table}[t]
\centering
\begin{adjustbox}{max width=0.48\textwidth}
\begin{tabular}{||c | c||} 
 \hline
\textbf{Base representations} & \textbf{Combinations} \\
 \hline \hline
 $\textbf{Only-x}\langle \vec{v_x} \rangle$ & $\textbf{Diff}  \oplus \textbf{Mult}$\\
 $\textbf{Only-y}\langle \vec{v_y} \rangle$ & $\textbf{Diff}  \oplus \textbf{Sqdiff}$\\
 $\textbf{Diff}\langle \vec{v_y} - \vec{v_x} \rangle$ & $\textbf{Sum}  \oplus \textbf{Mult}$\\
$\textbf{Sum}\langle \vec{v_x} + \vec{v_y} \rangle$ & $\textbf{Sum}  \oplus \textbf{Sqdiff}$\\
$\textbf{Concat}\langle \vec{v_x} \oplus \vec{v_y} \rangle$ &$\textbf{Concat}  \oplus \textbf{Mult}$ \\
$\textbf{Mult}\langle \vec{v_x} \odot \vec{v_y} \rangle$ & $\textbf{Concat}  \oplus \textbf{Sqdiff}$\\
$\textbf{Sqdiff}\langle (\vec{v_y} - \vec{v_x}) \odot  (\vec{v_y} - \vec{v_x}) \rangle$ &\\
\hline
\end{tabular}
\end{adjustbox}
\vspace*{-5pt}
\caption{Word pair representations.}
\label{table:1}
\vspace*{-7pt}
\end{table}
\begin{table*}[t]
\centering
\begin{adjustbox}{max width=0.98\textwidth}
\begin{tabular}{||c | c | c | c||} 
 \hline
\bfseries  Dataset & \bfseries Relations &  \bfseries \#Instances & \bfseries \#Domains\\
 \hline\hline
 \smalltt{\textbf{BLESS}} & attri (attribute), coord (co-hyponym), event, hyper (hypernymy), mero (meronymy), random&26,554&17\\
  \hline
\smalltt{\textbf{K\&H+N}}&hypo (hypernymy), mero (meronymy), sibl (co-hyponym), false (random)&63,718&3\\
 \hline
 \smalltt{\textbf{ROOT09}}&hyper (hypernymy), coord (co-hyponym), random &12,762&--\\
  \hline
\multirow{2}{*}{\smalltt{\textbf{EVALution}}}&HasProperty (attribute), synonym, HasA (possession), &\multirow{2}{*}{7,378}&\multirow{2}{*}{--}\\
& MadeOf (meronymy), IsA (hypernymy), antonym, PartOf (meronymy) & &\\
 \hline
\end{tabular}
\end{adjustbox}
\vspace*{-5pt}
\caption{Metadata on the datasets. Relations are mapped to corresponding WordNet relations, if available.}
\label{table:2}
\vspace*{-6pt}
\end{table*}

We classify each word pair $(x,y)$ to a specific semantic relation that holds for them, from a set of pre-defined relations (i.e., multiclass classification), based on their distributional representations.

\subsection{Word Pair Representations}
\label{sec:representation}
Given a word pair $(x,y)$ and their embeddings $\vec{v_x}, \vec{v_y}$, we consider various compositions as feature vectors for classifiers. Table \ref{table:1} displays base representations and combination representations, achieved by concatenating two base representations.

\subsection{Word Vectors}
We used $300$-dimensional pre-trained word embeddings, namely, \texttt{GloVe} \cite{Pennington:14} containing 1.9M word vectors trained on a corpus of web data from Common Crawl (42B tokens),\footnote{\tiny \url{http://nlp.stanford.edu/projects/glove/}} and \texttt{Word2vec} \cite{Mikolov:13a, Mikolov:13b} containing 3M word vectors trained on a part of Google News dataset (100B tokens).\footnote{\tiny \url{http://code.google.com/p/word2vec/}} Out-of-vocabulary words were initialized randomly. 

\subsection{Classifiers}
Following previous work \cite{Levy:15,Roller:16}, we trained different types of classifiers for each word-pair representation outlined in Section~\ref{sec:representation}, namely, logistic regression with $L_2$ regularization (\smalltt{LR}), SVM with a linear kernel (\smalltt{LIN}), and SVM with a Gaussian kernel (\smalltt{RBF}). In addition, we trained multi-layer perceptrons with a single hidden layer (\smalltt{MLP}). We compare our models against the \smalltt{KSIM} model found to be successful in previous work \cite{Levy:15,Kruszewski:15}. We do not include \newcite{Roller:16}'s model since it focuses only on hypernymy. Hyper-parameters are tuned using grid search, and we report the test performance of the hyper-parameters that performed best on the validation set. Below are more details about the training procedure:

\vspace*{-2mm}
\begin{itemize}
\item For \smalltt{LR}, the inverse of regularization strength is selected from $\{2^{-1}, 2^{1}, 2^{3}, 2^{5}\}$.
\vspace*{-1mm}
\item For \smalltt{LIN}, the penalty parameter $C$ of the error term is selected from $\{2^{-5}, 2^{-3}, 2^{-1}, 2^{1}\}$.
\vspace*{-3mm}
\item For \smalltt{RBF}, $C$ and $\gamma$ values are selected from $\{2^{1}, 2^{3}, 2^{5}, 2^{7}\}$ and $\{2^{-7}, 2^{-5}, 2^{-3}, 2^{-1}\}$, respectively.
\vspace*{-3mm}
\item For \smalltt{MLP}, the hidden layer size is either 50 or 100, and the learning rate is fixed at $10^{-3}$. We use early stopping based on the performance on the validation set. The maximum number of training epochs is 100.
\vspace*{-3mm}
\item For \smalltt{KSIM}, $C$ and $\alpha$ values are selected from $\{2^{-7}, 2^{-5}, \ldots, 2^{7}\}$ and $\{0.0, 0.1, \ldots, 1.0\}$, respectively.
\end{itemize}
\vspace*{-3mm}

\subsection{Datasets}
We evaluated the methods on four common semantic relation datasets: \smalltt{BLESS} \cite{Baroni:11}, \smalltt{K\&H+N} \cite{Necsulescu:15}, \smalltt{ROOT09} \cite{Santus:16}, and \smalltt{EVALution} \cite{Santus:15}. Table \ref{table:2} provides metadata on the datasets. Most datasets contain word pairs instantiating different, explicitly typed semantic relations, plus 
a number of unrelated word pairs (\textit{random}). Instances in \smalltt{BLESS} and \smalltt{K\&H+N} are divided into a number of topical domains.\footnote{We discarded two relations in \texttt{EVALution} with too few instances and did not include its domain information since each word pair can belong to multiple domains at once.}

\subsection{Evaluation Setup}
We consider the following evaluation setups:
\paragraph{Random (RAND)} We randomly split each dataset into 70\% train, 5\% validation and 25\% test.

\paragraph{Lexical Split (LEX)} In line with recent work \cite{Shwartz:16a}, we split each dataset into train, validation and test sets so that each contains a distinct vocabulary. This differs from \newcite{Levy:15} who dedicated a subset of the train set for evaluation, allowing the model to memorize when tuning hyper-parameters. We tried to keep the same ratio $70:5:25$ as in the random setup. 

\paragraph{Out-of-domain (OOD)} To test whether the methods capture a generic notion of each semantic relation, we test them on a domain that the classifiers have not seen during training. This setup is more realistic than the random and lexical split setups, in which the classifiers can benefit from memorizing verbatim words (random) or regions in the vector space (lexical split) that fit a specific slot of each relation.

Specifically, on \smalltt{BLESS} and \smalltt{K\&H+N}, one domain is held out for testing whilst the classifiers are trained and validated on the remaining domains. This process is repeated using each domain as the test set, and each time, a randomly selected domain among the remaining domains is left out for validation. The average results are reported.

\section{Experiments}
\label{sec:experiments}
\renewcommand{\arraystretch}{1.2}
\begin{table*}[ht]
%\hspace*{-20pt}
\centering
\scriptsize
\begin{adjustbox}{max width=\textwidth}
\begin{tabular}{||c||c||c|l|c|l|c||c|l|c|l|c||c||} 
 \hline
 
 \multirow{2}{*}{\textbf{Setup}} & \multirow{2}{*}{\textbf{Dataset}} &  \multicolumn{5}{|c||}{\textbf{Linear classifiers (LR, LIN)}}  & \multicolumn{5}{|c||}{\textbf{Non-linear classifiers (RBF, MLP)}} & \multirow{2}{*}{\texttt{\textbf{KSIM}}} \\ \cline{3-12}
  
& & $\vec{v_y}$ & \multicolumn{2}{|c|}{\textbf{Base}} & \multicolumn{2}{|c||}{\textbf{Combination}} & $\vec{v_y}$ & \multicolumn{2}{|c|}{\textbf{Base}} & \multicolumn{2}{|c||}{\textbf{Combination}} & \\
 
  \hline \hline 
  
% RAND
 \multirow{8}{*}{{\textbf{RAND}}} & \texttt{\textbf{BLESS}} & 84.4 & \specialcell{LR\\Concat} & 83.8 & \specialcell{LR\\Concat $\oplus$ Mult} & 89.5 (\textbf{+5.7}) & 89.3 & \specialcell{RBF\\Concat} & 94.0 & \specialcell{RBF\\Concat $\oplus$ Mult} & 94.3 (\textbf{+0.3}) & 70.2 \\ \cline{2-13}
& \texttt{\textbf{K\&H-N}} & 89.1 & \specialcell{LR\\Concat} & 95.4 & \specialcell{LR\\Concat $\oplus$ SqDiff} & 96.1 (\textbf{+0.7}) & 96.4 & \specialcell{RBF\\Concat}  & 98.6 & \specialcell{RBF\\Concat $\oplus$ Mult} & 98.6 (\textbf{0.0}) & 82.4 \\ \cline{2-13}
& \texttt{\textbf{ROOT09}} & 68.5 & \specialcell{LIN\\Sum} & 65.9 & \specialcell{LIN\\Sum $\oplus$ Mult} & 84.6 (\textbf{+18.7}) & 66.1 & \specialcell{RBF\\Sum} & 87.3 & \specialcell{RBF\\Sum $\oplus$ SqDiff} & 88.8 (\textbf{+1.5}) & 72.3 \\ \cline{2-13}
& \texttt{\textbf{EVALution}} & 49.7 & \specialcell{LIN\\Concat} & 56.7 & \specialcell{LIN\\Concat $\oplus$ Mult} & 56.8 (\textbf{+0.1}) & 52.1 & \specialcell{RBF\\Concat} & 61.1 & \specialcell{RBF\\Concat $\oplus$ Mult} & 60.6 (\textbf{-0.5}) & 50.5 \\ 

% LEX
 \hline \hline 
\multirow{8}{*}{{\textbf{LEX}}} & \texttt{\textbf{\textbf{BLESS}}} & 69.9 & \specialcell{LIN\\Concat} & 70.6 & \specialcell{LIN\\Concat $\oplus$ Mult} & 74.5 (\textbf{+3.9}) & 69.8 & \specialcell{MLP\\Concat} & 63.0 & \specialcell{MLP\\Concat $\oplus$ Mult} & 73.8 (\textbf{+10.8}) & 65.8 \\ \cline{2-13}
& \texttt{\textbf{K\&H-N}} & 78.3 & \specialcell{LIN\\Sum} & 74.0 & \specialcell{LIN\\Sum $\oplus$ SqDiff} & 76.1 (\textbf{+2.1}) & 83.2 & \specialcell{RBF\\Sum} & 82.0 & \specialcell{RBF\\Sum $\oplus$ Mult} & 81.7 (\textbf{-0.3}) & 77.5 \\ \cline{2-13}
& \texttt{\textbf{\textbf{ROOT09}}} & 66.7 & \specialcell{LR\\Concat} & 66.0 & \specialcell{LR\\Concat $\oplus$ Mult} & 77.9 (\textbf{+11.9}) & 64.5 & \specialcell{RBF\\Concat} & 76.8 & \specialcell{RBF\\Concat $\oplus$ Mult} & 81.6 (\textbf{+4.8}) & 66.7 \\ \cline{2-13}
& \texttt{\textbf{EVALution}} & 35.0 & \specialcell{LR\\Concat} & 37.9 & \specialcell{LR\\Concat $\oplus$ Mult} & 40.2 (\textbf{+2.3}) & 35.5 & \specialcell{RBF\\Concat} & 43.1 & \specialcell{RBF\\Concat $\oplus$ Mult} & 44.9 (\textbf{+1.8}) & 35.9 \\
\hline\hline

% OOD
\multirow{4}{*}{{\textbf{OOD}}} & \texttt{\textbf{BLESS}} & 70.9 & \specialcell{LIN\\Concat} & 69.9 & \specialcell{LIN\\Concat $\oplus$ Mult} & 77.0 (\textbf{+7.1}) & 69.9 & \specialcell{RBF\\Diff} & 78.7 & \specialcell{RBF\\Diff $\oplus$ Mult} & 81.5 (\textbf{+2.8}) & 57.8 \\ \cline{2-13}
& \texttt{\textbf{K\&H-N}} & 38.5 & \specialcell{LIN\\Concat} & 38.6 & \specialcell{LIN\\Concat $\oplus$ Mult} & 39.7 (\textbf{+1.1}) & 48.6 & \specialcell{MLP\\Sum} & 44.7 & \specialcell{MLP\\Sum $\oplus$ Mult} & 47.9 (\textbf{+3.2}) & 48.9 \\
\hline
\end{tabular}
\end{adjustbox}
\vspace*{-6pt}
\caption{Best test performance ($F_1$) across different datasets and evaluation setups, using \texttt{Glove}. The number in brackets indicates the performance gap between the best performing combination  and base representation setups.
}

\label{table:3}
\end{table*}

\setlength\tabcolsep{4pt}
\renewcommand{\arraystretch}{1.5}
\begin{table*}[h!]
%\hspace*{-20pt}
\centering
\scriptsize
\begin{adjustbox}{max width=\textwidth}
\begin{tabular}{||c|c||c|c|c|c|c|c|c||c|c|c|c|c|c|c||} 
\hline
\multicolumn{2}{||c||}{\textbf{Vector/}} & \multicolumn{7}{|c||}{\textbf{RAND}} & \multicolumn{7}{|c||}{\textbf{OOD}} \\ \cline{3-16}
\multicolumn{2}{||c||}{\textbf{Classifier}} & $\vec{v_y}$ & \textbf{Diff} & \textbf{Diff} $\oplus$ \textbf{Mult} & \textbf{Sum} & \textbf{Sum} $\oplus$ \textbf{Mult} & \textbf{Concat} & \textbf{Concat} $\oplus$ \textbf{Mult} & $\vec{v_y}$ & \textbf{Diff} & \textbf{Diff} $\oplus$ \textbf{Mult} & \textbf{Sum} & \textbf{Sum} $\oplus$ \textbf{Mult} &
 \textbf{Concat} & \textbf{Concat} $\oplus$ \textbf{Mult} \\
 
 % Glove
 \hline \hline
\parbox[t]{1.0mm}{\multirow{4}{*}{\rotatebox[origin=c]{90}{GloVe}}} & \textbf{LR} & 84.4 & 81.5 & 87.6 \textbf{(\textbf{+6.1})}& 81.5 & 87.0 (\textbf{+5.5}) & 83.8 & 89.5 (\textbf{+5.7}) & 70.9 & 64.5 & 74.7 (\textbf{+10.2}) & 59.2 &
68.9 (\textbf{+9.7}) & 69.5 & 76.5 (\textbf{+7.0})\\
& \textbf{LIN} & 84.1 & 81.5 & 87.7 \textbf{(\textbf{+6.2})}& 81.3 & 87.2 (\textbf{+5.9}) & 83.8 & 89.2 (\textbf{+5.4}) & 70.7 & 64.6 & 74.8 (\textbf{+10.2}) & 59.3 &
69.4 (\textbf{+10.1}) & 69.9 & 77.0 (\textbf{+7.1})\\
& \textbf{RBF} & 89.3 & 93.8 & 94.1 \textbf{(\textbf{+0.3})}& 94.4 & 94.2 (\textbf{\textbf{-0.2}}) & 94.0 & \textbf{94.3} (\textbf{+0.3}) & 67.8 & 78.7 & \textbf{81.5} (\textbf{+2.8}) & 65.3 &
66.4 (\textbf{+1.1}) & 69.5 & 75.7 (\textbf{+6.2}) \\
& \textbf{MLP} & 84.4 & 87.4 & 89.2 \textbf{(\textbf{+1.8})}& 87.2 & 89.9 (\textbf{+2.7}) & 90.5 & 90.5 (\textbf{0.0}) & 69.9 & 67.4 & 77.7 (\textbf{+10.3}) & 57.3 & 
66.1 (\textbf{+8.8}) & 71.5 & 77.3 (\textbf{+5.8}) \\

% Word2vec
 \hline \hline
\parbox[t]{1.0mm}{\multirow{4}{*}{\rotatebox[origin=c]{90}{Word2vec}}} &
\textbf{LR} & 83.5 & 81.0 & 85.4 (\textbf{+4.4}) & 80.0 & 84.6 (\textbf{+4.6}) & 83.6 & 87.1 (\textbf{+3.5}) & 71.2 & 62.4 & 69.0 (\textbf{+6.6}) & 59.0 & 65.3 (\textbf{+6.3})
& 71.8 & \textbf{76.1} (\textbf{+4.3}) \\
& \textbf{LIN} & 83.3 & 80.8  & 84.6 (\textbf{+3.8}) & 80.4 & 84.5 (\textbf{+4.1}) & 83.3 & 86.5 (\textbf{+3.2}) & 71.5 & 62.8 & 69.1 (\textbf{+6.3}) & 59.8 & 65.2 (\textbf{+5.4})
& 72.1 & 76.0 (\textbf{+3.9})\\
& \textbf{RBF} & 89.1 & 93.7 & 93.7 (\textbf{0.0}) & 93.7 & \textbf{93.8} (\textbf{+0.1}) & 93.6 & \textbf{93.8} (\textbf{+0.2}) & 69.2 & 75.6 & 76.0 (\textbf{+0.4}) & 64.7 & 66.3 (\textbf{+1.6}) & 71.4 & 75.3 (\textbf{+3.9})\\
& \textbf{MLP} & 81.6 & 81.0 & 84.6 (\textbf{+3.6}) & 79.6 & 85.2 (\textbf{+5.6}) & 81.3 & 84.7 (\textbf{+3.4}) & 70.2 & 63.4 & 69.3 (\textbf{+5.9}) & 56.2 & 60.0 (\textbf{+3.8})
& 70.5 &  74.6 (\textbf{+4.1})\\
\hline
\end{tabular}
\end{adjustbox}
\vspace*{-6pt}
\caption{Test performance ($F_1$) on \texttt{BLESS} in the RAND and OOD setups, using \texttt{Glove} and \texttt{Word2vec}.}
\label{table:4}
\end{table*}

Table~\ref{table:3} summarizes the best performing base representations and combinations on the test sets across the various datasets and evaluation setups.\footnote{Due to the space limitation, we only show the results obtained with \texttt{Glove}. The trend is similar across the word embeddings.} The results across the datasets vary substantially in some cases due to the differences between the datasets' relations, class balance, and the source from which they were created. For instance, \smalltt{K\&H+N} is imbalanced between the number of instances across relations and domains. \smalltt{ROOT09} was designed to mitigate the lexical memorization issue by adding negative switched hyponym-hypernym pairs to the dataset, making it an inherently more difficult dataset. \smalltt{EVALution} contains a richer set of semantic relations. Overall, the addition of multiplicative features improves upon the performance of the base representations.

\paragraph{Classifiers}
Multiplicative features substantially boost the performance of linear classifiers. However, the gain from adding multiplicative features is smaller when non-linear classifiers are used, since they partially capture such notion of interaction \cite{Levy:15}. Within the same representation, there is a clear preference to non-linear classifiers over linear classifiers.

\paragraph{Evaluation Setup}
The \textbf{Only-y} representation indicates how well a model can perform without considering the relation between $x$ and $y$ \cite{Levy:15}. Indeed, in RAND, this method performs similarly to the others, except on \smalltt{ROOT09}, which by design disables lexical memorization. As expected, a general decrease in performance is observed in LEX and OOD, stemming from the methods' inability to benefit from lexical memorization. In these setups, there is a more significant gain from using multiplicative features when non-linear classifiers are used.

\paragraph{Word Pair Representations} Among the base representations \textbf{Concat} often performed best, while \textbf{Mult} seemed to be the preferred multiplicative addition. \textbf{Concat} $\oplus$ \textbf{Mult} performed consistently well, intuitively because \textbf{Concat} captures the typicality of each word in the relation (e.g., if $y$ is a typical hypernym) and \textbf{Mult} captures the similarity between the words (where \textbf{Concat} alone may suggest that \textit{animal} is a hypernym of \textit{apple}). To take a closer look at the gain from adding \textbf{Mult}, Table~\ref{table:4} shows the performance of the various base representations and combinations with \textbf{Mult} using different classifiers on \smalltt{BLESS}.\footnote{We also tried $\vec{v_x}$ with multiplicative features but they performed worse.}

\section{Analysis of Multiplicative Features}
\label{sec:analysis}
\setlength\tabcolsep{6pt}
\renewcommand{\arraystretch}{1.0}
\begin{table*}[t]
\centering
\small
\begin{tabular}{||c c c c c c||} 
 \hline
$x$ & \textbf{relation} & $y$ & \textbf{similarity} & \textbf{Concat} & \textbf{Concat} $\oplus$ \textbf{Mult} \\
 \hline\hline
 cloak-n 	& random & good-j & 0.195 & attribute &	random \\
 cloak-n 	& random & hurl-v & 0.161 & event & random \\
 cloak-n 	& random & stop-v & 0.186 & event & random \\
 coat-n & event & wear-v & 0.544 & random & event \\
 cloak-n & mero & silk-n & 0.381 & random & mero \\
 dress-n & attri & feminine-j & 0.479 &	random & attri \\
 \hline
\end{tabular}
\vspace*{-4pt}
\caption{Example pairs which were incorrectly classified by \textbf{Concat} while being correctly classified by \textbf{Concat} $\oplus$ \textbf{Mult} in \smalltt{BLESS}, along with their cosine similarity scores.}
\label{table:5}
\end{table*}

We focus the rest of the discussion on the OOD setup, as we believe it is the most challenging setup, forcing methods to consider the relation between $x$ and $y$. We found that in this setup, all methods performed poorly on \smalltt{K\&H+N}, likely due to its imbalanced domain and relation distribution. Examining the per-relation $F_1$ scores, we see that many methods classify all pairs to one relation. Even \smalltt{KSIM}, the best performing method in this setup, classifies pairs as either \emph{hyper} or \emph{random}, effectively only determining if they are related or not. We therefore focus our analysis on \smalltt{BLESS}.

To get a better intuition of the contribution of multiplicative features, Table~\ref{table:5} exemplifies pairs that were incorrectly classified by \textbf{Concat} (\smalltt{RBF}) while correctly classified by \textbf{Concat} $\oplus$ \textbf{Mult} (\smalltt{RBF}), along with their cosine similarity scores. It seems that \textbf{Mult} indeed captures the similarity between $x$ and $y$. While \textbf{Concat} sometimes relies on properties of a single word, e.g. classifying an adjective $y$ to the \emph{attribute} relation and a verb $y$ to the \emph{event} relation, adding \textbf{Mult} changes the classification of such pairs with low similarity scores to \emph{random}. Conversely, pairs with high similarity scores which were falsely classified as \emph{random} by \textbf{Concat} are assigned specific relations by\\ \textbf{Concat} $\oplus$ \textbf{Mult}. 

Interestingly, we found that across domains, there is an almost consistent order of relations with respect to mean intra-pair cosine similarity: 

\setlength\tabcolsep{3pt}
\begin{table}[h]
\centering
\small
\begin{adjustbox}{max width=0.48\textwidth}
\begin{tabular}{||c c c c c c||} 
 \hline
\textbf{coord} & \textbf{meronym}  & \textbf{attribute} & \textbf{event} & \textbf{hypernym} & \textbf{random} \\
 \hline\hline
 0.426 & 0.323	& 0.304	& 0.296	& 0.279	& 0.141\\
 \hline
\end{tabular}
\end{adjustbox}
\vspace{-10pt}
\caption{Mean pairwise cosine similarity in \texttt{BLESS}.}
\label{table:6}
\end{table}

Since the difference between \emph{random} (0.141) and other relations (0.279-0.426) was the most significant, it seems that multiplicative features help distinguishing between related and unrelated pairs. 
This similarity is possibly also used to distinguish between other relations.

\section{Conclusion}
\label{sec:conclusion}
We have suggested a cheap way to boost the performance of supervised distributional methods for lexical entailment by integrating multiplicative features into standard word-pair representations. Our results confirm that the multiplicative features boost the performance of linear classifiers, and in strict evaluation setups, also of non-linear classifiers. We performed an extensive evaluation with different classifiers and evaluation setups, and suggest the out-of-domain evaluation as the most suitable for the task. Directions for future work include investigating other compositions, and designing a neural model that can automatically learn such features.

\section{Acknowledgements}

We would like to thank Wei Lu for his involvement and advice in the early stage of this project, Stephen Roller and Omer Levy for valuable discussions, and the anonymous reviewers for their insightful comments and suggestions.

Vered is supported in part by an Intel ICRI-CI grant, the Israel Science Foundation grant 1951/17, the German Research Foundation through the German-Israeli Project Cooperation (DIP, grant DA 1600/1-1), the Clore Scholars Programme (2017), and the AI2 Key Scientific Challenges Program (2017).

\bibliography{references}
\bibliographystyle{acl_natbib}

\end{document}